\documentclass[letterpaper]{article} 
\usepackage[preprint]{aaai2027}  
\usepackage[hyphens]{url}  
\usepackage{graphicx} 
\usepackage{natbib}  
\usepackage{caption} 
\usepackage{algorithm}
\usepackage{algorithmic}
\usepackage{amsmath}
\usepackage{pifont}

\DeclareMathOperator*{\argtopk}{arg\,topk}

\newcommand{\taona}{T\textsc{ao}N\textsc{a}}
\newcommand{\taonaA}{\taona-A}
\usepackage{newfloat}
\usepackage{listings}
\DeclareCaptionStyle{ruled}{labelfont=normalfont,labelsep=colon,strut=off} 
\floatstyle{ruled}
\newfloat{listing}{tb}{lst}{}
\floatname{listing}{Listing}

\usepackage{booktabs}

\usepackage{tcolorbox}
\tcbuselibrary{skins,listings,breakable}

\newtcblisting{promptbox}[1]{
  enhanced,
  listing engine=listings,
  listing only,
  breakable,
  title={#1},
  colback=black!5,
  colframe=black!75,
  colbacktitle=black!75,
  coltitle=white,
  arc=2mm,
  boxrule=0.8pt,
  fonttitle=\bfseries\normalsize,
  boxsep=0.5mm,
  left=1.5mm,
  right=1.5mm,
  top=1mm,
  bottom=1mm,
  toptitle=1mm,
  bottomtitle=1mm,
  listing options={
    basicstyle=\rmfamily\footnotesize,
    numbers=none,
    breaklines=true,
    breakatwhitespace=false,
    columns=fullflexible,
    keepspaces=true,
    showstringspaces=false,
    xleftmargin=0pt,
    aboveskip=0pt,
    belowskip=0pt,
    breakindent=0pt,
    breakautoindent=false
  }
}

\title{Towards a Learnable Retriever for Textual Graph Question Answering}
\author{
    Kaifeng Jin\textsuperscript{\rm 1},
    Yuchen Yan\textsuperscript{\rm 1},
    Zhichen Zeng\textsuperscript{\rm 1},
    Xiao Lin\textsuperscript{\rm 1},
    Lihui Liu\textsuperscript{\rm 2},
    Yinglong Xia\textsuperscript{\rm 3},
    Hanghang Tong\textsuperscript{\rm 1}
}

\affiliations{
    \textsuperscript{\rm 1}University of Illinois Urbana-Champaign\\
    \textsuperscript{\rm 2}Wayne State University\\
    \textsuperscript{\rm 3}Meta
}

\newcommand{\algname}{\textsc{TaLenT}}

\newcommand{\cmark}{\ding{51}}
\newcommand{\xmark}{\ding{55}}

\DeclareMathOperator*{\argmax}{arg\,max}

\newcommand{\G}{\mathcal{G}}
\newcommand{\V}{\mathcal{V}}
\newcommand{\E}{\mathcal{E}}

\begin{document} 

\maketitle

\begin{abstract} Textual graph question answering (GraphQA) and graph-based retrieval-augmented generation (GraphRAG) have gained increasing attention as a way to ground large language models (LLMs) on structured knowledge. Despite extensive efforts, existing methods still face two key gaps. The first is an \textit{architectural gap}: existing methods typically fine-tune an LLM-based generator, but do not also make the retriever itself a learnable LLM for graph evidence selection within one end-to-end retrieve-then-generate pipeline. The second is a \textit{retrieval-supervision gap}: existing retrievers rarely use gold answer signals to supervise evidence selection.
As a result, they may fail to recall evidence nodes required for reasoning, resulting in insufficient \textit{coverage} of the gold answer set; meanwhile, naive attempts to boost recall often introduce excessive irrelevant information that compromises \textit{compactness} and overwhelms the LLM. To address these gaps, we propose \algname, an LLM-based end-to-end retriever-generator framework for textual GraphQA. To bridge the architectural gap, \algname\ pairs a learnable LLM-based retriever for evidence selection with a fine-tuned LLM-based generator within one unified retrieve-then-generate pipeline. To bridge the retrieval-supervision gap, we leverage gold answer signals as retriever supervision.
Specifically, we stratify graph nodes into gold answers, answer-related nodes, and irrelevant noise. We then adopt a weighted pseudo-label loss to prioritize gold answers and preserve answer-related nodes for coverage, while discarding irrelevant noise for compactness. Experimental results across three GraphQA datasets demonstrate that \algname\ consistently improves downstream performance. \end{abstract}

\section{Introduction}

\begin{figure*}[t]
  \centering
  \includegraphics[width=\textwidth]{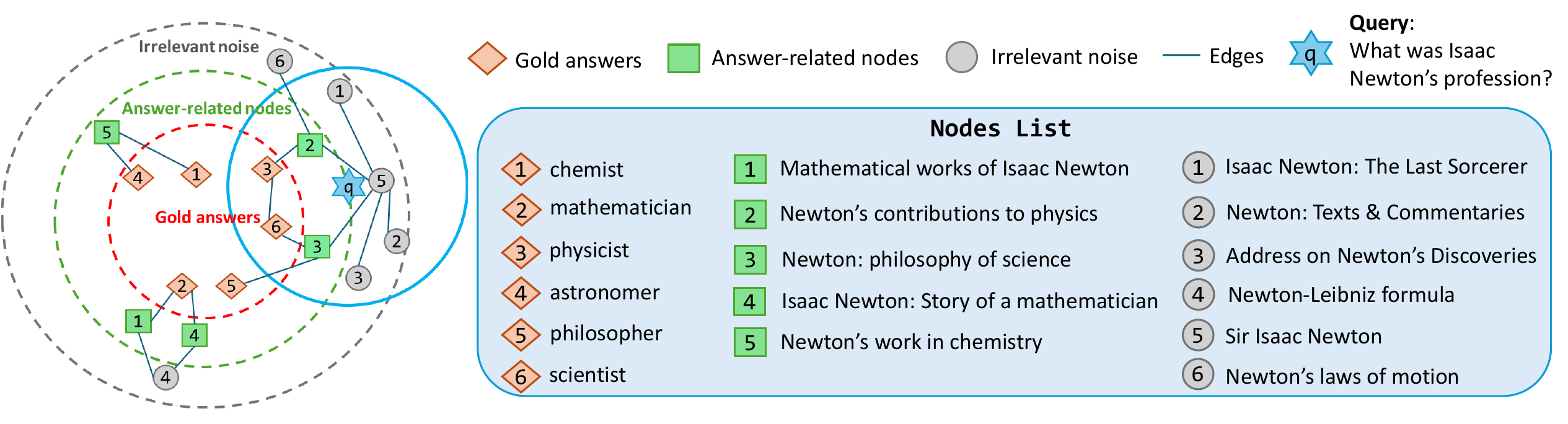}
  \caption{An illustrative example of two limitations of query-based graph retrieval.
  For the query ``What was Isaac Newton's profession?'', a query-similarity retriever returns a Newton-centric subgraph (blue circle). We partition the nodes into three groups: \emph{Gold answers} (only a small subset of the gold answer set is retrieved by the subgraph in the blue circle, showing limited \textit{coverage}), \emph{Answer-related nodes} (topically relevant but not exact gold answers), and \emph{Irrelevant noise} (nodes that match the query tokens such as ``Newton'' yet are not answer-related). The presence of substantial noise alongside partial gold coverage highlights the \textit{compactness} issue: improving recall by expanding query-similar candidates can inflate the subgraph with distractors.}
  \label{fig:teaser}
\end{figure*}

Large language models (LLMs) have achieved impressive progress in general-purpose generation and reasoning \cite{thoppilan2022lamda, bang2023multitask}, but their outputs are bounded by what is encoded in model parameters.
Trained on large-scale open-domain corpora collected before a fixed cutoff date, LLMs may hallucinate or omit critical evidence when a query requires precise, up-to-date, or domain-specific knowledge \cite{zhao2023survey, wang2023knowledge}.

To mitigate these limitations, Retrieval-Augmented Generation (RAG) has emerged as a dominant paradigm that grounds LLM outputs in external evidence by dynamically retrieving relevant knowledge snippets for context-aware generation \cite{lewis2020retrieval, shi2024replug, gao2023precise}. While conventional RAG methods \cite{lewis2020retrieval, shi2024replug} mainly retrieve evidence from unstructured text corpora, recent work has shifted toward textual graphs as a more expressive and structured knowledge source. This trend has spurred the development of Graph-based RAG (GraphRAG) and Graph Question Answering (GraphQA). By exploiting the explicit semantics and relational structure of textual graphs, graph-based RAG methods support more grounded reasoning and complex information synthesis \cite{logan2019barack, luo2025chatrule, he2024g}.

A graph-based RAG pipeline generally consists of two stages: graph evidence retrieval and answer generation.
On the generation side, existing methods often use an LLM-based generator and fine-tune it over retrieved graph contexts.
This makes graph evidence retrieval a critical step, since the generator can only reason over the evidence selected for it.

Conventional retrieval methods in GraphRAG can be categorized into non-parametric, graph neural network (GNN)-based, and LM-based approaches.
Non-parametric retrievers construct subgraphs using graph search algorithms or predefined rules \cite{taunk2023grapeqa, yasunaga2021qa}.
GNN-based methods encode graph structure with GNNs and identify task-relevant components to form the retrieved subgraph \cite{mavromatis2024gnn}.
LM-based methods often use LMs to assist graph retrieval, for example by reading structured data, generating retrieval cues, or using LM reasoning to guide evidence selection \cite{baek2023knowledge, jiang2023structgpt}.

However, with these retrieval designs, existing graph-based RAG pipelines still leave an architectural gap.
While the generator is often fine-tuned as an LLM-based module, the retriever itself is usually not made a learnable LLM for graph evidence selection.
Instead, evidence selection is still handled by non-parametric, GNN-based, or LM-assisted procedures.
As a result, the retriever and generator are not both trained as LLM modules within one end-to-end retrieve-then-generate pipeline.

Beyond this architectural gap, these conventional retrievers also face a retrieval-supervision gap.
They rely primarily on query similarity while rarely using gold
answers to supervise evidence selection.
Query similarity can favor nodes matching query terms over
answer-critical evidence, leading to two limitations: \textit{coverage}, where a query-based retriever may fail to cover the gold answer set in the retrieved subgraph, and \textit{compactness}, where it may introduce excessive irrelevant noise.
An illustrative example is shown in Figure~\ref{fig:teaser}.
A key observation is that the \emph{gold answer set} provides precisely the missing supervision for retrieval.
Gold answers specify what should be covered and also help distinguish answer-related evidence from irrelevant noise, and can therefore help alleviate these two limitations.
However, such gold signals are rarely exploited to guide or train graph retrievers.

In this work, we address both gaps by proposing \algname, an LLM-based end-to-end retriever-generator
framework for textual GraphQA.
To bridge the architectural gap, \algname\ makes the retriever itself a learnable LLM for graph evidence selection, and pairs it with a fine-tuned LLM-based generator within one unified retrieve-then-generate pipeline. Within the full pipeline, we first apply a rule-based
preprocessing step to construct a candidate subgraph, providing a high-recall pool of nodes under the LLM context budget.
The learnable LLM-based retriever is then trained to select relevant evidence nodes, and the generator is fine-tuned on contexts constructed from the trained retriever's outputs. To bridge the retrieval-supervision gap, we use gold answer signals as retriever supervision to improve both retrieval coverage and compactness. To improve retrieval coverage, we train a retriever based on a weighted pseudo-label objective, incorporating gold answer signals to guide the retrieval process. To promote answer-critical evidence, we prioritize gold answer and down-weight
other rest signals to improve retrieval compactness. Intuitively, the loss assigns the highest weight to gold answers to ensure coverage, assigns a much smaller weight to answer-related nodes to retain useful context, and ignores irrelevant noise to enforce compactness. In this way, \algname\ improves evidence selection before generation, leading to better downstream QA performance. Our contributions are summarized as follows:

\begin{itemize}
  \item \textbf{Idea.} We introduce an LLM-based end-to-end retriever-generator framework for textual GraphQA, where a learnable LLM-based retriever and a fine-tuned LLM-based generator are placed within one unified retrieve-then-generate pipeline. This design fills the architectural gap by making the retriever itself a learnable LLM for
graph evidence selection, rather than only fine-tuning the downstream generator.

  \item \textbf{Method.} We train the learnable LLM-based retriever with a weighted pseudo-label objective that leverages gold answer signals. By prioritizing gold answer nodes and down-weighting weakly related nodes, the objective fills the retrieval-supervision gap and leads to better downstream QA performance.

  \item \textbf{Evaluation.} We conduct extensive experiments on three GraphQA benchmarks to validate both retrieval effectiveness and end-task QA performance. Our method outperforms the best competitor by at least 3.18\%.
\end{itemize}

\section{Preliminaries}
\label{sec:prelim}

\noindent\textbf{Notation.}
We use calligraphic letters to denote sets, e.g., $\mathcal{V}$ and $\mathcal{E}$ for node and edge sets.
A textual graph is represented as $\mathcal{G}=(\mathcal{V},\mathcal{E})$, where each node $v\in\mathcal{V}$ has a textual attribute $\mathrm{text}(v)$, e.g., an entity name or concept, and each edge $e=(u,r,v)\in\mathcal{E}$ denotes a relation $r$ connecting the source node $u$ to the destination node $v$. The textual attribute of an edge is denoted as $\mathrm{text}(e)$.
We use $\mathcal{S}$ to denote the subgraph of $\mathcal{G}$, where $\mathcal{S}\subseteq \mathcal{G}$ indicates that its node set $\mathcal{V}_\mathcal{S} \subseteq \mathcal{V}$ and its edge set $\mathcal{E}_\mathcal{S} \subseteq \mathcal{E}$. We use $q$ for the query and $a$ for the final answer from the generator. We use $a^{\star}$ to denote the token sequence of the gold answer in the dataset and use $\mathcal{V}^{\star}$ as the set of nodes appearing in the gold answer, i.e., the gold answer set. We denote the retriever and generator by $R_{\theta}$ and $G_{\phi}$, respectively.

\subsection{Problem Setup}
\label{subsec:problem_setup}

We consider a general retrieve-then-generate setup \cite{he2024g} for textual graph question answering (GraphQA).
Each instance in the dataset consists of a natural-language query $q$ and a textual graph $\mathcal{G}=(\mathcal{V},\mathcal{E})$ as the knowledge base.
Due to the large scale of $\mathcal{G}$, it is impractical to directly serialize $\G$ into an LLM prompt, hence the system first retrieves a relevant \emph{subgraph}, conditioned on which an answer is generated.

Formally, a retriever $R_\theta$ extracts a subgraph $\mathcal{S}$ from the textual graph $\mathcal{G}$ based on the query $q$:
\begin{equation}
\mathcal{S} = R_\theta(q,\mathcal{G}).
\end{equation}

A generator $G_\phi$ then predicts the answer $a$ based on the query $q$ and the retrieved subgraph $\mathcal{S}$:

\begin{equation}
a = G_\phi(q,\mathcal{S}).
\end{equation}

\subsection{Candidate Connected Subgraph}
\label{subsec:pcst_prelim}

For large textual graphs, directly serializing the full graph often exceeds the LLM context budget. As in prior graph-based QA pipelines, we include a preprocessing stage. Specifically, following G-Retriever~\cite{he2024g}, we first construct a \emph{candidate connected subgraph} ${\mathcal{S}}$ using the Prize-Collecting Steiner Tree (PCST) algorithm~\cite{bienstock1993note}, and use $\mathcal{S}$ as the input context for downstream modules.

\noindent\textbf{Indexing and similarity retrieval.}
Given a query $q$ and a textual graph $\mathcal{G}=(\mathcal{V},\mathcal{E})$, we embed the query and textual attributes of node $\mathcal{V}$ and edges $\mathcal{E}$ using a sentence encoder such as SentenceBERT~\cite{reimers2019sentence}:

\begin{align}
\mathbf{z}_q &= \mathrm{LM}(q), \\
\mathbf{z}_v &= \mathrm{LM}(\mathrm{text}(v)), \quad v\in \mathcal{V}, \\
\mathbf{z}_e &= \mathrm{LM}(\mathrm{text}(e)), \quad e\in \mathcal{E}.
\end{align}

We compute cosine similarity scores and retrieve the top-$k_v$ nodes and top-$k_e$ edges:

\begin{align}
\mathcal{V}_{\text{sim}}
&= \argtopk_{v\in\mathcal{V},\,k=k_v}\ \cos(\mathbf{z}_q,\mathbf{z}_v), \\
\mathcal{E}_{\text{sim}}
&= \argtopk_{e\in\mathcal{E},\,k=k_e}\ \cos(\mathbf{z}_q,\mathbf{z}_e),
\end{align}

where $k_v$ and $k_e$ control the number of retrieved nodes and edges, respectively.

\noindent\textbf{Subgraph construction.}
We assign \emph{prizes} to retrieved candidates based on their ranks, so higher-ranked items receive larger prizes.
Concretely, for $v\in \mathcal{V}_{\text{sim}}$ with rank $r_v$ we set $p(v)=k_v-r_v$ to prioritize those with higher similarity, and analogously define $p(e)=k_e-r_e$ for $e\in \mathcal{E}_{\text{sim}}$; items outside these retrieved lists receive zero prize.
PCST then selects a connected subgraph $\mathcal{S}=(\V_{\mathcal{S}},\E_{\mathcal{S}})$ that trades off prize coverage and compactness:

\begin{equation}
\label{eq:pcst_obj}
\begin{aligned}
\mathcal{S}
= \argmax_{\substack{\mathcal{S}\subseteq \mathcal{G}\\ \text{connected}}}
\Big(
\sum_{v\in \V_{\mathcal{S}}} p(v)
+ \sum_{e\in \E_{\mathcal{S}}} p(e)
- c\,|\mathcal{E}_\mathcal{S}|
\Big).
\end{aligned}
\end{equation}

where $c$ controls the size-coverage trade-off.

\section{Method}
\label{sec:method}

\begin{figure*}
    \centering
    \includegraphics[width=0.9\textwidth]{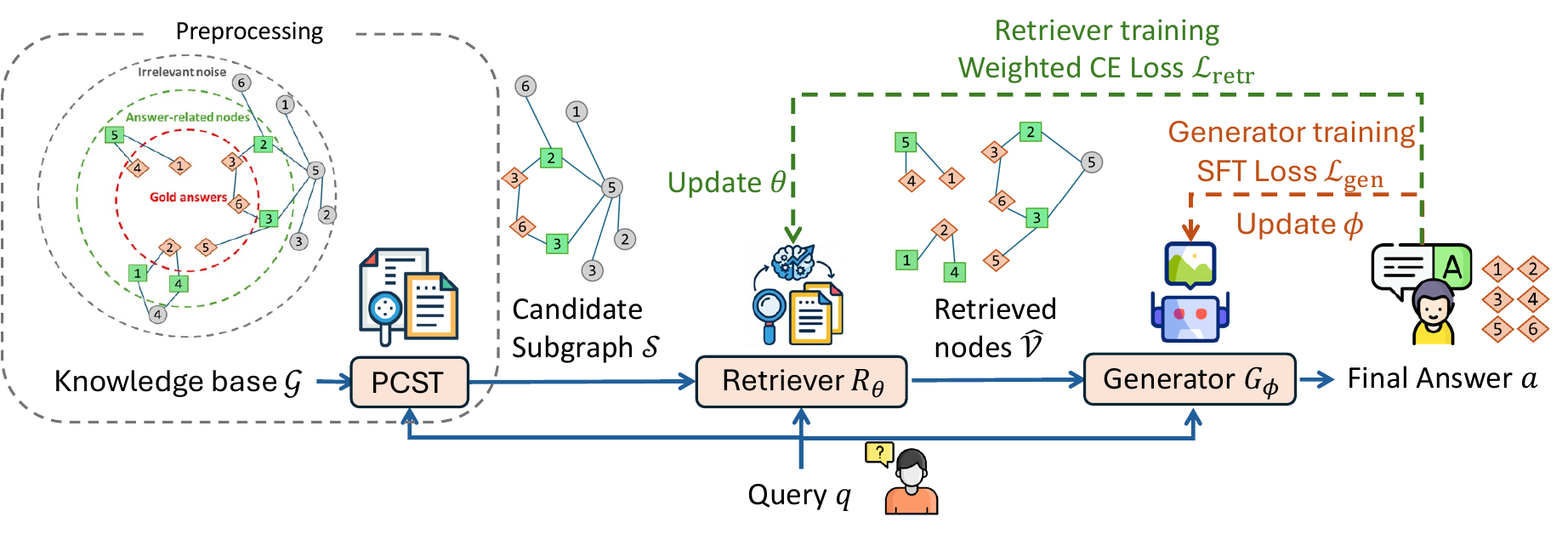}
    \caption{Overview of \algname, an LLM-based end-to-end retriever-generator framework for textual GraphQA. The full \algname\ pipeline contains three stages: (1) Preprocessing, which applies PCST to construct an initial candidate connected subgraph $\mathcal{S}$ from the knowledge base $\mathcal{G}$ based on query $q$. (2) Retriever training, which trains a learnable LLM-based retriever $R_{\theta}$ with weighted pseudo-label supervision, given the candidate subgraph $\mathcal{S}$ and query $q$, to produce $\hat{\mathcal{V}}$. (3) Generator training, which fine-tunes an LLM-based generator $G_{\phi}$ with standard SFT given retrieved node set $\hat{\mathcal{V}}$ and $q$, to produce the final answer $a$.}
    \label{fig:overview}
\end{figure*}
\noindent\textbf{Overview.}
\algname\ is an LLM-based end-to-end retriever-generator framework for textual GraphQA.
Its two core trainable LLM modules are:
(i) a learnable nodes retriever $R_{\theta}$ trained with weighted pseudo-label supervision (Section \ref{subsec:retriever_training}), and 
(ii) an answer generator $G_{\phi}$ fine-tuned with standard Supervised Fine-Tuning (SFT) (Section \ref{subsec:generator_ft}). The full \algname\ pipeline is illustrated in Figure~\ref{fig:overview}. Given a query $q$ and a full textual graph $\mathcal{G}$, a rule-based PCST preprocessing step first constructs an initial candidate connected subgraph $\mathcal{S}$ that fits the LLM context budget.
We then train a learnable retriever $R_{\theta}$ with weighted pseudo-label supervision to produce an answer-relevant node set $\hat{\mathcal{V}}$ (Section \ref{subsec:retriever_training}).
Using answer-relevant nodes $\hat{\mathcal{V}}$, we construct textual contexts and fine-tune a generator $G_{\phi}$ with standard supervised learning (Section \ref{subsec:generator_ft}).

\subsection{Retriever Training}
\label{subsec:retriever_training}

As discussed earlier, PCST and other query-similarity based retrievers often fail to cover the answer-related nodes and include irrelevant noise, resulting in limited \emph{coverage} and poor \emph{compactness}. We therefore use gold answer signals to guide retriever training, encouraging prioritization of gold answers while down-weighting other signals. We first describe \algname's retriever prompting and then introduce the weighted pseudo-label objective.

\noindent\textbf{Retriever prompting.}
We implement our learnable retriever $R_{\theta}$ as an instruction-following LLM (with LoRA) that performs node selection based on a provided candidate subgraph.
Given a query $q$ and the PCST subgraph $\mathcal{S}$, we prompt the LLM to produce the nodes that are most helpful for answering $q$. The LLM outputs should be exactly one line in the format of ``\texttt{name1|name2|...}''. Each \texttt{name} such as \texttt{name1} and \texttt{name2} represents the names of the nodes $R_{\theta}$ finally selects. 
This constrained format makes the output directly interpretable as a retrieved nodes set, which we denote by
$\hat{\mathcal{V}} = R_\theta(q, {\mathcal{S}})$. A concrete retriever prompt example is provided in Appendix~\ref{app:retriever-prompt}. It is worth noting that since our retriever $R_\theta$ is an LLM, it leverages its inherent parametric knowledge during the selection process. Consequently, the output node set $\hat{\mathcal{V}}$ may include nodes that were not present in the initial candidate subgraph $\mathcal{S}$ (i.e., $\hat{\mathcal{V}} \not\subseteq \mathcal{V}_\mathcal{S}$). In our method, $\hat{\mathcal{V}}$ encompasses the complete set of nodes generated by $R_\theta$, reflecting both the information from $\mathcal{S}$ and the LLM's internal knowledge.

\noindent\textbf{Fine-tuning with weighted pseudo-label supervision.}
Recall that existing works in GraphRAG use retrievers that are largely query-similarity driven, which leads to two limitations: coverage and compactness.
Our goal is to train the retriever to output the node set $\hat{\mathcal{V}}$ that achieves both high \emph{coverage} and \emph{compactness}:
it should reliably include nodes that appear in the gold answer set $\mathcal{V}^{\star}$ (coverage) while avoiding redundant or weakly related candidates (compactness). The gold answer provides precisely the missing supervision, i.e., the gold signal, for retrieval.
To inject gold signals into retrieval, we leverage the dataset-provided gold answers and stratify candidate nodes into three levels:
(i) \emph{gold answers}, which are candidate nodes that overlap with the gold answer set $\mathcal{V}^{\star}$, and provides evidence that the retriever should prioritize to support correct reasoning,  (ii) \emph{answer-related nodes}, which are candidate nodes within $\mathcal{S}$ but are not gold answers, and may still be helpful for providing context, though they are less directly tied to the final answer,
and (iii) \emph{irrelevant noise}, which are nodes not contained in $\mathcal{S}$,  and are completely irrelevant and should be suppressed to maintain compactness. Given the candidate subgraph $\mathcal{S}$ with node set $\V_{\mathcal{S}}$, and the gold answer set $\mathcal{V}^{\star}$, we utilize these nodes as weighted pseudo label supervision by defining

\begin{equation}
    \mathcal{P} = \mathcal{V}^{\star} \cap \V_{\mathcal{S}} \quad \text{(strong positives; level-1)},
\end{equation}
\begin{equation}
    \mathcal{W} = \V_{\mathcal{S}} \setminus \mathcal{V}^{\star} \quad \text{(weak positives; level-2)}.
\end{equation}
All nodes outside $\V_{\mathcal{S}}$ are treated as irrelevant noise (level-3) during retriever training.

We train $R_{\theta}$ as conditional generation of a target string $y$ that concatenates nodes in $\mathcal{P}$ and $\mathcal{W}$ (separated by the pipe symbol ``|''; strong positives first, then weak positives)

\begin{equation}
y = p_1 \mid \cdots \mid p_{|\mathcal{P}|} \mid w_1 \mid \cdots \mid w_{|\mathcal{W}|},
\end{equation}
where $p_i, \forall i=1,\dots|\mathcal{P}|$ are nodes in $\mathcal{P}$ and $w_j, \forall j=1\dots|\mathcal{W}|$ are nodes in $\mathcal{W}$.
Let $x$ denote the retriever input prompt, which includes the query $q$ and the textualized candidate subgraph $\mathcal{S}$, and let $y_{1:T}$ be the tokenized target sequence, where $T$ is its token length.
To prioritize level-1 (gold answer) nodes while still preserving level-2 (answer-related) candidates as a weak learning signal, we minimize the token-weighted negative log-likelihood:

\begin{equation}
\mathcal{L}_{\text{retr}}(\theta)
= - \sum_{t=1}^{T} \omega_t \log p_{\theta}(y_t \mid x, y_{<t}).
\label{eq:weighted_ce}
\end{equation}

Since we want to encourage $R_{\theta}$ to prioritize gold answer nodes while still allowing potentially useful answer-related nodes to contribute as a weak learning signal, we adopt the following token weights to achieve this

\begin{equation}
\omega_t =
\begin{cases}
1, & \text{if } y_t \text{ belongs to a node in } \mathcal{P},\\
\lambda, & \text{if } y_t \text{ belongs to a node in } \mathcal{W},\\
0, & \text{otherwise,}
\end{cases}
\end{equation}
where $\lambda$ is a hyperparameter that controls the contribution of weak positives relative to strong positives.

\begin{table*}[t]
\centering
{\small
\begin{tabular}{lccc}
\hline
Dataset (Metrics) & ExplaGraphs (ACC) & WebQSP (Hit@1) & CWQ (Hit@1) \\
\hline
Zero-shot & 56.50 & 41.06 & 18.58 \\
Zero-CoT \cite{kojima2022large} & 57.04 & 51.30 & 25.63 \\
CoT-BAG \cite{wang2023can} & 57.94 & 39.60 & 18.83 \\
KAPING \cite{baek2023knowledge} & 62.27 & 52.64 & 23.25 \\
Graph-based Inference & 33.93 & 47.22 & 18.21 \\
Frozen LLM + Prompt Tuning (PT) & 57.63 & 48.34 & 36.22 \\
GraphToken \cite{perozzi2024letgraphtalkingencoding} & 85.08 & 57.05 & 33.39 \\
G-Retriever & 85.16 & 70.49 & 45.62 \\
G-Retriever w/ LoRA & \underline{87.05} & \underline{73.79} & \underline{49.19} \\
\taonaA~\cite{yan2025answer} & 87.01 & 71.23 & 45.28 \\
StructGPT \cite{jiang2023structgpt} & 78.34 & 72.60 & 32.57 \\
\hline
\textbf{\algname} & \textbf{91.79} & \textbf{76.97} & \textbf{55.71} \\
\hline
\end{tabular}
}
\caption{Benchmark results (\%).}
\label{tab:main_results}
\end{table*}
\subsection{Generator Training}
\label{subsec:generator_ft}

\noindent\textbf{Generator Prompting.} Given the predicted node set $\hat{\mathcal{V}} = R_\theta(q, \mathcal{S})$ output by the trained $R_\theta$, we form the generator input as a \emph{prompt} that concatenates the query $q$ with subgraphs derived from $\hat{\mathcal{V}}$.
These subgraphs contain (i) the retrieved nodes in $\hat{\mathcal{V}}$ and (ii) their associated edges $\hat{\mathcal{E}}$ that appear in $\mathcal{S}$.
Concretely, we include edges for which at least one endpoint is in $\hat{\mathcal{V}}$:

\begin{equation}
\hat{\mathcal{E}}
=
\{(u,r,v)\in\mathcal{E}: u\in\hat{\mathcal{V}} \ \text{or}\ v\in\hat{\mathcal{V}}\}.
\end{equation}
We then serialize the retrieved nodes $\hat{\mathcal{V}}$ together with the selected edges $\hat{\mathcal{E}}$ into a \emph{serialized graph text} and prepend the query $q$ to form the generator prompt. Importantly, this \emph{serialized graph text} is not required to form one whole connected subgraph. This design avoids forcing unnecessary ``bridge'' nodes or edges solely to maintain connectivity, which often increases irrelevant context and hurts compactness.
Instead, we allow the retrieved evidence to appear as multiple disconnected components, as long as it covers the nodes in $\hat{\mathcal{V}}$.
Moreover, as we mentioned in Section \ref{subsec:retriever_training}, $\hat{\mathcal{V}}$ may include nodes that come from $R_\theta$'s parametric knowledge and do not appear in the initial candidate subgraph $\mathcal{S}$, we retain these nodes as additional isolated nodes in the \emph{serialized graph text}. A concrete generator prompt example together with the graph serialization format is provided in Appendix~\ref{app:generator-prompt}.

\noindent\textbf{Fine-tuning with standard SFT.}
We fine-tune an LLM-based generator $G_{\phi}$ using standard supervised fine-tuning (SFT).
For each training instance, we first run the trained retriever $R_\theta$ to obtain the generator prompt constructed in the previous paragraph (i.e., the \emph{serialized graph text} formed from $\hat{\mathcal{V}}$ and the associated $\hat{\mathcal{E}}$, together with the query $q$). We then train $G_{\phi}$ to generate the gold answer $a^{\star}$ conditioned on this prompt:

\begin{equation}
\mathcal{L}_{\text{gen}}(\phi)
= - \sum_{t=1}^{T_a} \log p_{\phi}(a^{\star}_t \mid q, \hat{\mathcal{V}}, \hat{\mathcal{E}}, a^{\star}_{<t}),
\end{equation}
where $a^{\star}=(a^{\star}_1,\ldots,a^{\star}_{T_a})$ denotes the \emph{token} sequence of the gold answer, and $T_a$ is its length.
This SFT step transfers retriever's improved evidence selection into downstream QA.

\subsection{Inference and Evaluation}
\label{subsec:inference}

At inference time, we apply the full retrieve-then-generate pipeline using the retriever and generator fine-tuned on the training (and validation) splits.
For each test instance $(q,\mathcal{G})$, we first apply the fixed
PCST-based preprocessing step to construct the candidate subgraph $\mathcal{S}$, then run the learned retriever to obtain $\hat{\mathcal{V}}=R_{\theta}(q,\mathcal{S})$.
We materialize $(\hat{\mathcal{V}},\hat{\mathcal{E}})$ into a textual graph context and feed it to the generator to produce the final answer $a=G_{\phi}(q,\hat{\mathcal{V}},\hat{\mathcal{E}})$.
We report task metrics computed from the generator outputs on the test split.

\section{Experiments}
\label{sec:results}

\subsection{Experiment Setup}

\noindent\textbf{Datasets and Metrics.} To evaluate our method on textual QA tasks, we conduct experiments on three textual graph question answering datasets: ExplaGraphs and WebQSP from the GraphQA benchmark~\cite{he2024g}, and CWQ~\cite{talmor2018web}. Specifically,
(1) ExplaGraphs's task is to determine whether two arguments are supportive or contradictory. 
(2) WebQSP is a knowledge graph question answering dataset whose primary task is to answer questions that require multi-hop reasoning. (3) CWQ is a more compositional KGQA benchmark than WebQSP, with more complex questions and longer reasoning chains. For evaluation, we adopt Accuracy for ExplaGraphs and Hit@1 for WebQSP and CWQ. Dataset statistics and additional dataset details are provided in Appendix~\ref{app:datasets}. 

\noindent\textbf{Baselines.} We compare against two groups of baselines. (1) Inference-Only methods that include Zero-shot, Zero-CoT~\cite{kojima2022large}, CoT-BAG~\cite{wang2023can}, KAPING~\cite{baek2023knowledge}, Graph-based Inference, and StructGPT~\cite{jiang2023structgpt}. (2) Prompt-Tuning methods that include Frozen LLM + Prompt Tuning (PT), GraphToken~\cite{perozzi2024letgraphtalkingencoding}, G-Retriever~\cite{he2024g}, and \taonaA~\cite{yan2025answer}. For G-Retriever, we include a variant that fine-tunes the LLM-based generator with LoRA (G-Retriever w/ LoRA). Additional discussion of our differences from other answer-aware or answer-related methods and recent graph retrieval baselines not included in Table~\ref{tab:main_results} is provided in Appendix~\ref{app:additional-baselines}.

\noindent\textbf{Experiment pipeline.} For the main experiments, we use the open-source LLaMA2-Chat-7B model~\citep{touvron2023llama} as the backbone for both the retriever and the generator, and conduct all experiments on a single NVIDIA A100 GPU. We follow the same train/validation/test splits as prior work for each dataset. For WebQSP and CWQ, since the full KG exceeds the context budget, we first construct a PCST candidate subgraph with $k_v{=}k_e{=}10$ and then apply our learnable retriever. During retriever training, we use the weighted pseudo-label objective with strong-positive weight $1$. The weak-positive weight is set to $\lambda=0.001$ for WebQSP and
$\lambda=0.01$ for ExplaGraphs and CWQ. Additional implementation details, including hyperparameter
configurations and computing infrastructure, are provided in Appendix~\ref{app:hyperparams}. Appendix~\ref{app:sensitivity} reports sensitivity analyses for $\lambda$ and PCST size parameters $k_v$ and $k_e$. A backbone ablation with Llama-3-8B~\citep{grattafiori2024llama} is provided in Appendix~\ref{app:llama3_ablation}.

\subsection{Benchmark results}
\label{subsec:main_results}

\begin{figure*}[t]
\centering

\includegraphics[width=0.95\textwidth]{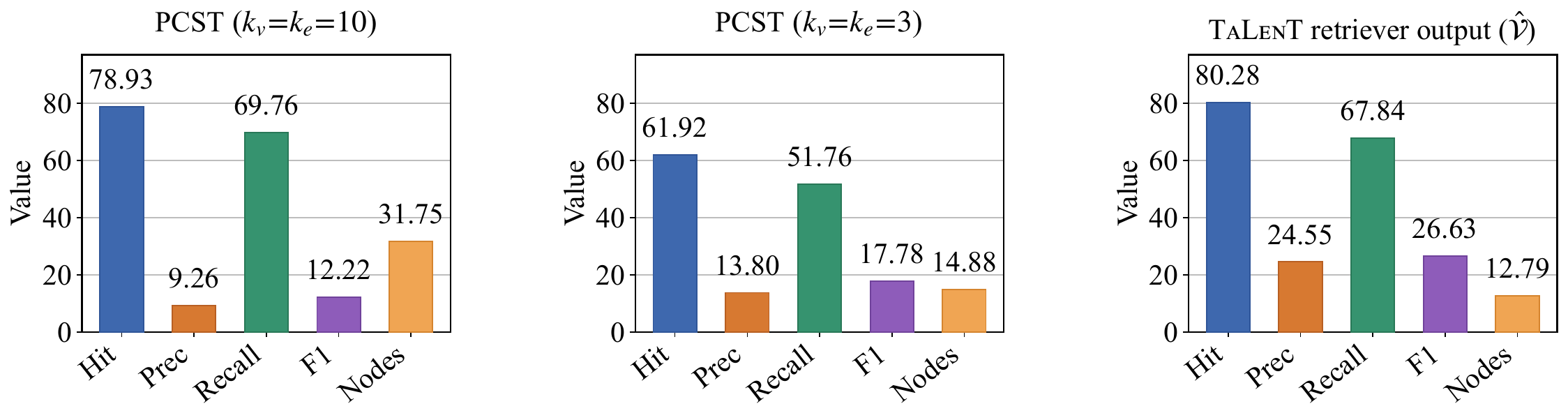}

\caption{WebQSP test-set retrieval results (macro-averaged). 
We measure how well PCST subgraphs and \algname's learned retriever output $\hat{\mathcal{V}}$ cover the gold answer set $\mathcal{V}^{\star}$.
We compare: (1) the PCST candidate subgraph with $k_v{=}k_e{=}10$ (this is the retriever input), 
(2) a much smaller PCST subgraph with $k_v{=}k_e{=}3$ (a size-matched reference), 
and (3) the nodes set $\hat{\mathcal{V}}$ selected by \algname's learned retriever.
Our retriever keeps coverage close to PCST($k_v{=}k_e{=}10$) while using far fewer nodes, and it also covers more gold nodes than the size-matched PCST($k_v{=}k_e{=}3$).}

\label{fig:retr_efficiency_bar}
\end{figure*}

Table~\ref{tab:main_results} reports the benchmark results on ExplaGraphs (Accuracy), WebQSP (Hit@1), and CWQ (Hit@1).
Overall, \algname\ outperforms the strongest competitors across all three datasets. We draw the following three observations.

\noindent\textbf{(1) \algname\ outperforms the strongest competitors.}
\algname\ achieves 91.79\% accuracy on ExplaGraphs, 76.97\% Hit@1 on WebQSP, and 55.71\% Hit@1 on CWQ, improving over the strongest
baseline by 4.74\% on ExplaGraphs, 3.18\% on WebQSP, and 6.52\% on CWQ.
These results show that \algname\ generalizes well under different graph QA objectives and metrics, with the largest gain obtained on the more challenging CWQ benchmark.

\noindent\textbf{(2) Comparison to inference-only methods.}
Inference-only baselines remain behind \algname\ on all three datasets.
Even compared with the strongest inference-only baseline, StructGPT, \algname\ improves by 13.45\% on ExplaGraphs, 4.37\% on WebQSP, and 23.14\% on CWQ.
This performance gap stems from their reliance on test-time prompting and non-trainable evidence acquisition, without task-specific fine-tuning of either the retriever or generator. In contrast, \algname\ trains both components, enabling more reliable evidence selection and more accurate answers.

\noindent\textbf{(3) Comparison to prompt-tuning methods.}
Prompt-tuning methods form substantially stronger baselines than inference-only methods overall, with G-Retriever w/ LoRA being the strongest prior prompt-tuning baseline on all three datasets.
Nevertheless, \algname\ consistently improves over it, from 87.05\% to 91.79\% on ExplaGraphs, from 73.79\% to 76.97\% on WebQSP, and from 49.19\% to 55.71\% on CWQ.
These gains support the two central design choices of \algname. First, \algname\ moves beyond rule-based, query-driven evidence selection by introducing a learnable LLM-based retriever for task-adaptive graph evidence selection within an end-to-end retriever-generator framework.
Second, it trains this retriever with gold-answer-guided weighted pseudo-label supervision to improve evidence coverage and compactness, thereby benefiting downstream generation.

\subsection{Coverage-Compactness Tradeoff}
\label{subsec:retr_tradeoff}

In this section, we study how \algname’s retriever achieves a better coverage-compactness tradeoff for its output node set $\hat{\mathcal{V}}$. We conduct experiments on WebQSP, with results shown in Figure~\ref{fig:retr_efficiency_bar}, and make the following observations.

\noindent\textbf{Compactness: similar coverage with substantially fewer nodes.}
We first compare the compactness of nodes retrieved by the initial PCST candidate subgraph ($k_v{=}k_e{=}10$) and \algname, as shown in Figure~\ref{fig:retr_efficiency_bar}. Under similar gold-node coverage, \algname’s retriever output $\hat{\mathcal{V}}$ contains significantly fewer nodes. Both methods achieve comparable Hit@1 and Recall on WebQSP, yet $\hat{\mathcal{V}}$ contains less than half as many nodes: on average 12.79 nodes for \algname~ versus 31.75 nodes for the PCST subgraph. Meanwhile, $\hat{\mathcal{V}}$ attains notably higher precision and F1, improving precision from 9.26\% to 24.55\% and F1 from 12.22\% to 26.63\%. These results suggest that the weighted pseudo-label supervision effectively incorporates gold signals, enabling the retriever to filter redundant or weakly related nodes and produce a more compact context without sacrificing coverage.

\noindent\textbf{Coverage: better gold answer coverage under comparable node number.}
To further compare coverage, we report results for a much smaller PCST subgraph with $k_v{=}k_e{=}3$, whose size is comparable to that of \algname. As shown in Figure~\ref{fig:retr_efficiency_bar}, under a similar node budget, \algname’s retriever achieves substantially higher coverage. Specifically, \algname\ improves Hit@1 from 61.92\% to 80.28\% and Recall from 51.76\% to 67.84\% compared to PCST. This demonstrates that, given a similar number of nodes, our gold-signal-involved weighted pseudo-label supervision enables the retriever to achieve significantly higher coverage than the rule-based, query-similarity-driven PCST algorithm.

\begin{table}[t]
\centering
\begin{tabular}{lcc}
\toprule
\textbf{Output subset} & \textbf{Hit@1}  & \textbf{Recall} \\
\midrule
All retrieved nodes & 80.28 & 67.84  \\
Nodes within PCST & 75.31 & 64.19 \\
Nodes out of PCST & 10.26 & 5.46 \\
\bottomrule
\end{tabular}
\caption{Breakdown of gold answer coverage contributed by different parts of the learned retriever output on WebQSP test. 
All metrics are macro-averaged over examples and reported as percentages.}
\label{tab:retr_breakdown}
\end{table}

{\setlength{\tabcolsep}{1mm}
\begin{table}[t]
\centering
\begin{tabular}{lcccc}
\toprule
\textbf{Retriever setting} & \textbf{Hit@1} & \textbf{Prec} & \textbf{Recall} & \textbf{F1}  \\
\midrule
\algname\ & 80.28 & 24.55 & 67.84 & 26.63 \\
Ablate retriever training & 36.61 & 14.97 & 26.26 & 14.97 \\
\bottomrule
\end{tabular}
\caption{Retriever ablation on WebQSP test set (macro-averaged over examples). ``Ablate retriever training'' replaces \algname's learnable retriever by a frozen retriever without LoRA fine-tuning.}
\label{tab:ablation_retrieval}
\end{table}
}

\begin{table}[t]
\centering
{\small
\begin{tabular}{lc}
\toprule
\textbf{System} & \textbf{Hit@1 (\%)} \\
\midrule
\algname\ w/o retriever training & 68.37 \\
\algname\ w/o retriever & 72.50 \\
\algname\ & 76.97  \\
\bottomrule
\end{tabular}}
\caption{QA ablations on WebQSP (Hit@1, \%). All systems follow the same pipeline as \algname\ except for the ablated component stated in each row.}
\label{tab:ablation_qa}
\end{table}

\subsection{Retrieval Beyond PCST}

Fine-tuning on an LLM-based retriever may leverage parametric knowledge to retrieve nodes beyond the PCST candidate subgraph. We therefore examine whether \algname’s retriever can retrieve such beyond-PCST nodes and how they contribute to performance. Specifically, we split the retriever output $\hat{\mathcal{V}}$ into nodes inside and outside the PCST subgraph. Shown in Table~\ref{tab:retr_breakdown}, nodes outside PCST alone recover a non-trivial portion of gold answer nodes, with a Hit@1 of 10.26\% and a Recall of 5.46\%.  Moreover, incorporating beyond-PCST nodes improves Hit@1 from $75.31$ to $80.28\%$ and Recall from $64.19\%$ to $67.84\%$. Results indicate that the trained LLM retriever effectively complements the PCST candidate subgraph with parametric-knowledge nodes that enhance gold-node coverage. An additional groundedness and hallucination analysis of out-of-PCST retrieval is provided in Appendix~\ref{app:groundedness-hallucination}.
\section{Ablation Study}
\label{sec:ablation}

We conduct two ablation studies on WebQSP to illustrate (i) the effect of pseudo-label retriever training (ii) the effect of learnable retriever.

\subsection{On the Effect of Pseudo-label Training}
\label{subsec:ablation_pseudo}

\noindent\textbf{Retrieval quality.}
Table~\ref{tab:ablation_retrieval} compares \algname's retriever against an ablated variant, which uses the same backbone LLaMA2-Chat-7B model but keeps it \emph{frozen} (no training) and directly prompts it to select helpful nodes from the PCST candidate subgraph.
Results show that the frozen retriever fails to recover gold nodes required for answering, with Hit@1, Precision, and Recall of 36.61\%, 14.97\%, and 26.26\%, respectively.
In contrast, \algname's retriever explicitly prioritizes gold answer nodes during retriever fine-tuning using pseudo label training, substantially improving coverage with a Hit@1 of 80.28\%, a precision of 24.55\% and a recall of 67.84\%, indicating that injecting gold signals into retriever training is crucial for selecting answer-relevant nodes.

\noindent\textbf{Downstream QA performance.}
We next examine whether the retrieval improvement translates into better end-task QA.
Table~\ref{tab:ablation_qa} summarizes QA Hit@1 under three systems: (i) \algname\ with a \emph{frozen} retriever (ablate training on retriever), (ii) \algname\ \emph{without} the retriever module (directly feeding the PCST candidate subgraph to the generator), and (iii) the full \algname\ system.
Switching from a frozen retriever to our pseudo-label trained retriever improves QA Hit@1 from 68.37\% to 76.97\%.
This gain is consistent with Table~\ref{tab:ablation_retrieval}: pseudo-label training recovers substantially more gold nodes, providing the generator with more reliable evidence.

\subsection{On the Effect of Learnable Retriever}
\label{subsec:ablation_module}
We further isolate the contribution of the retriever by removing it from the pipeline and directly feeding the PCST candidate subgraph into the generator, i.e., without $R_{\theta}$ or $\hat{\mathcal{V}}$ selection. As shown in Table~\ref{tab:ablation_qa}, removing the learnable retriever reduces QA performance from 76.97\% to 72.50\%, indicating that an explicit retriever provides clear benefits for downstream QA. Notably, using a frozen retriever performs even worse (Hit@1 of 68.37\%) than removing the retriever entirely, suggesting that retrieval is not inherently beneficial when gold answer node coverage is low, as noisy context can mislead the generator. This result further underscores the importance of our weighted pseudo-label training, which substantially improves retrieval quality for end-task QA.

\section{Related Work}
\label{sec:related_work}

\noindent\textbf{Retrieval Augmented Generation.}
RAG grounds LLMs with external evidence to reduce hallucinations and improve faithfulness \cite{sun2024product, gao2023precise}.
Classic RAG systems follow a straightforward index-retrieve-generate workflow \cite{ma2023query}, while many recent variants strengthen retrieval with additional processing before and after retrieval.
For example, pre-retrieval techniques such as query transformation, expansion, and rewriting are often adopted to better match the search space \cite{zheng2023take, gao2023precise}, and post-retrieval reranking is commonly used to refine the evidence set \cite{blagojevi2023enhancing}.
More broadly, modular RAG systems use LLMs to iteratively refine retrieval queries or evidence selection \cite{yu2022generate}.

\noindent\textbf{RAG over Graphs.}
Extensive works study how to combine graph-structured information with LLMs \cite{li2023survey, wang2023can}.
Prior efforts span general graph modeling \cite{liu2023one, lei2023boosting}, multimodal architectures that incorporate graph-like structures \cite{yoon2023multimodal, li2023graphadapter}.
Representative applications include fundamental graph reasoning \cite{zhao2023graphtext}, node prediction \cite{sun2023large, chen2024exploring}, graph-level prediction \cite{qian2023can}.
A related line of work discusses GraphRAG / GraphQA, where the system selects graph evidence and conditions an LLM on graph context to answer questions \cite{logan2019barack,luo2025chatrule}.

\section{Conclusion}
\label{sec:conclusion}
We address two key gaps in existing GraphRAG systems: an \textit{architectural gap} and a \textit{retrieval-supervision gap}. To address both gaps, we propose \algname, an LLM-based end-to-end retriever-generator framework for textual GraphQA. To bridge the architectural gap, \algname\ makes the retriever itself a learnable LLM for graph evidence selection and pairs it with a fine-tuned LLM-based generator within one unified retrieve-then-generate pipeline. To bridge the retrieval-supervision gap, we train \algname's retriever with a gold-answer-guided weighted pseudo-label loss that prioritizes answer-related signals to improve coverage while suppressing noise for compactness. Extensive experiments on three GraphQA benchmarks demonstrate that
\algname\ consistently improves downstream QA performance.

\bibliography{reference}

\appendix

\section{Additional Dataset Details}
\label{app:datasets}

\subsection{Datasets}

\noindent\textbf{Datasets and Metrics.} To evaluate the effectiveness of our method on textual QA tasks, we conduct experiments on three textual graph question answering datasets, ExplaGraphs and WebQSP from the GraphQA benchmark~\cite{he2024g}, and CWQ~\cite{talmor2018web}. Specifically,
(1) ExplaGraphs is developed for generative commonsense reasoning, with a specific focus on the construction of explanation graphs for stance prediction in debates. By providing detailed and unambiguous commonsense-augmented graphs, it enables an evaluation of whether arguments either support or refute a specific belief. ExplaGraphs's primary task is to determine whether the arguments are supportive or contradictory. Its node denotes commonsense concepts and the edge denotes Commonsense relations. In this dataset, the gold answer is a binary label (i.e., `support' or `counter'), which does not provide explicit supervision for the retriever. Therefore, we leverage the LLaMA2-Chat-7B model \cite{touvron2023llama} to generate pseudo gold reasoning node sets, and treat these annotated nodes as pseudo-label supervision $\mathcal{V}^\star$ for training the learnable retriever. Due to the small graph size of the graphs in ExplaGraphs, we directly use the entire graph as the candidate context for retrieval. Following GraphQA~\cite{he2024g}, we evaluate the end task with \textbf{Accuracy}.
(2) WebQSP is a large-scale multi-hop knowledge graph QA dataset that consists of 4,737 questions. The primary task is to answer questions that necessitate multi-hop reasoning. It employs a subset of Freebase \cite{bollacker2008freebase}, with a specific focus on facts within 2 hops of the entities identified in the questions. The nodes denote Entities in Freebase and the edges denote Relations in Freebase. The gold answers are lists of Freebase entities, which naturally correspond to node lists and can directly serve as $\mathcal{V}^\star$ without requiring additional annotation. For metrics, we report Hit@1, counting a prediction as correct if the generated answer matches any of the answers in the gold answer list. (3)
CWQ (ComplexWebQuestions) is a compositional KGQA dataset constructed from WebQSP by generating more complex queries and rewriting them into natural language~\cite{talmor2018web}. It contains 34,689 questions. Compared with WebQSP, CWQ contains more complex questions and longer reasoning chains. Like WebQSP, CWQ is grounded in Freebase, where nodes denote entities and edges denote relations. The gold answers are lists of Freebase entities and can therefore directly serve as $\mathcal{V}^{\star}$. We evaluate CWQ with \textbf{Hit@1}. Dataset statistics are summarized in Table~\ref{tab:dataset_stats}. See Appendix \ref{app:explagraphs_annotation} for Pseudo-Gold Nodes Annotation for ExplaGraphs.

\begin{table*}[htbp]
\centering
\begin{tabular}{lccc}
\hline
Dataset & ExplaGraphs & WebQSP & CWQ\\
\hline
\#Graphs & 2,766 & 4,737 & 34,689 \\
Average \#Nodes & 5.17 & 1370.89 & 1255.71 \\
Average \#Edges & 4.25 & 4252.37 & 4000.86 \\
Node Attribute & Commonsense concepts & Entities in Freebase & Entities in Freebase \\
Edge Attribute & Commonsense relations & Relations in Freebase & Relations in Freebase  \\
Task & Commonsense reasoning & KGQA & KGQA\\
Evaluation metrics & Accuracy & Hit@1 & Hit@1\\
\hline
\end{tabular}
\caption{Statistics of datasets.}
\label{tab:dataset_stats}
\end{table*}

\subsection{Pseudo-Gold Nodes Annotation for ExplaGraphs}
\label{app:explagraphs_annotation}
ExplaGraphs provides gold supervision only as a stance label (support/counter), rather than node-level evidence annotations.
To make our learnable retriever pseudo label supervision applicable to this dataset, we introduce a lightweight preprocessing step on the training and validation splits:
we use \texttt{Llama-2-7b-chat} to annotate a set of pseudo-gold reasoning nodes for each example.
Concretely, we provide the model with (i) the full textual explanation graph, (ii) the two arguments, and (iii) the gold stance label, and prompt it to select nodes from the graph that form the critical reasoning path for deciding whether the arguments support or counter each other.
The selected nodes are then used as $\mathcal{V}^{\star}$ in our weighted pseudo-label objective for training the retriever.
This annotation is used only to train our retriever: the generator is still fine-tuned with the original task supervision (predicting support vs.\ counter), and at inference time the pipeline does not require access to any node annotations.

We further conduct two sanity checks on these pseudo-gold node annotations.
First, we manually inspected LLM annotations and found these LLM-selected nodes to be broadly consistent with human judgment.
For example, for the argument pair \textit{``Women should be allowed in military combat.''} and \textit{``Females as soldiers should be acceptable.''}, whose gold label is \textit{support}, the annotator selects critical nodes including \textit{strength}, \textit{military combat}, and \textit{females} as “pseudo-gold” nodes,  which we found to be reasonable and aligned with the human interpretation of key evidence needed to justify the support relation.

Second, we perform a cross-annotator consistency check using \texttt{GPT-4o-mini}.
Specifically, we re-annotate the ExplaGraphs dataset with \texttt{GPT-4o-mini} using the same annotation prompt as used for \texttt{Llama-2-7b-chat}, and compare the resulting node sets.
We report two simple agreement statistics.
(1) \textbf{Any-overlap rate}: in $80.44\%$ of graph instances, the two annotators share at least one selected node.
(2) \textbf{Reference coverage}: when treating the original \texttt{Llama-2-7b-chat} annotations as the reference, the \texttt{GPT-4o-mini} annotations cover $83.75\%$ of those nodes on average.
These results suggest that the pseudo-gold node annotations are not arbitrary: a second annotator recovers most of their core evidence, while manual inspection further indicates that the selected nodes are reasonable for stance prediction.
These sanity checks suggest that the \texttt{Llama-2-7b-chat} pseudo-gold annotations are reasonably consistent and provide useful supervision for training the ExplaGraphs retriever.

\section{Implementation Details}
\label{app:hyperparams}

\noindent\textbf{Preprocessing: Candidate-subgraph construction.}
For WebQSP and CWQ, we construct the PCST candidate subgraph with
$k_v{=}k_e{=}10$. Because the graphs in ExplaGraphs are small, we directly use the full graph as the candidate context.

\noindent\textbf{Retriever fine-tuning.}
We fine-tune the retriever with LoRA under our token-weighted cross-entropy objective.
Given a question and a candidate subgraph, we construct weighted pseudo labels over candidate nodes: nodes overlapping with gold answer nodes are treated as strong positives with weight $1.0$, while the remaining candidate nodes are treated as weak positives. The weak-positive weight is set to $\lambda=0.001$ for WebQSP and $\lambda=0.01$ for ExplaGraphs and CWQ. The training target is a single-line, pipe-separated node list, formed by concatenating strong positives and weak positives using the delimiter ``\texttt{|}'' (strong positives first).
We deduplicate node names in both retriever outputs and the contexts later used for generator training. 

We use LoRA with rank $r{=}8$, scaling factor $\alpha{=}16$, and dropout $0.05$.
Retriever training uses 3 epochs, learning rate $5\times10^{-5}$, warmup ratio $0.03$, per-device batch size 1 with gradient accumulation 8.

\noindent\textbf{Generator fine-tuning.}
We fine-tune the generator with LoRA using standard supervised fine-tuning on retrieved textual graph contexts and gold targets.
We use the same LoRA configuration ($r{=}8$, $\alpha{=}16$, dropout $0.05$) for generator fine-tuning.

\noindent\textbf{Computing infrastructure.}
The experiments were run on a server equipped with eight
NVIDIA A100-SXM4 GPUs, each with 80 GB of memory.
Each experimental run used a single GPU.
The server contained two Intel Xeon Gold 6448Y processors
(64 physical cores and 128 hardware threads in total) and 1 TiB of
system memory, and ran Ubuntu 22.04.5 LTS.
We used Python 3.10.19, NVIDIA driver 535.183.06,
PyTorch 2.5.1 with CUDA runtime 12.1 and cuDNN 9.1,
Transformers 4.57.1, PEFT 0.17.1, Accelerate 1.11.0, and NumPy 2.1.2.

\section{Sensitivity Analysis and Ablation of the Weighted Objective}
\label{app:sensitivity}

We provide two complementary analyses on WebQSP.
First, we study the sensitivity of our method to the weak-positive weight $\lambda$ and the PCST candidate subgraph size parameters $k_v$ and $k_e$.
In the $\lambda$ sweep, the setting $\lambda=1$ also serves as an
equal-weight ablation, since strong and weak positives receive the same
training weight.

\paragraph{Sensitivity to the weak-positive weight $\lambda$.}
We vary $\lambda$ while fixing the PCST input to $k_v = k_e = 10$.
Table~\ref{tab:sens_lambda} reports both downstream QA performance (Hit@1) and the average number of retrieved nodes on the test set, where a smaller number indicates better compactness.
Across all tested values of $\lambda$, downstream Hit@1 ranges from
$75.55\%$ to $76.97\%$, consistently exceeding the strongest baseline
reported in the main paper, G-Retriever w/ LoRA ($73.79\%$).
Our default setting, $\lambda=0.001$, achieves the highest Hit@1 of
$76.97\%$ while retaining a compact output of $12.79$ nodes on average.
At the same time, the average number of retrieved nodes increases consistently as $\lambda$ becomes larger, which is expected since larger weak-positive weights encourage the retriever to retain more weakly related candidate nodes. As $\lambda$ increases beyond 0.001, the retrieved node
sets become progressively larger without improving downstream performance. In particular, the equal-weight setting $\lambda=1$ reduces Hit@1 to $75.55\%$ and increases the average number of retrieved nodes from $12.79$ to $25.25$. These results support assigning weak positives a substantially smaller weight than strong positives, rather than weighting the two groups equally.

\begin{table}[t]
\centering
\begin{tabular}{ccc}
\hline
$\lambda$ & Hit@1 (\%) & Avg.\ \#Retrieved Nodes \\
\hline
0 & 76.23 & 10.19 \\
1e-5  & 76.66 & 10.38 \\
0.001  & 76.97 & 12.79 \\
0.1   & 76.23 & 22.42 \\
1.0   & 75.55 & 25.25 \\
\hline
\end{tabular}
\caption{Sensitivity analysis on the weak-positive weight $\lambda$ on WebQSP, with PCST input fixed to $k_v = k_e = 10$. We report Hit@1 and the average number of retrieved nodes on the test set; smaller retrieved node sets indicate better compactness. Larger $\lambda$ values lead to less compact retrieval.}
\label{tab:sens_lambda}
\end{table}

\paragraph{Sensitivity to PCST candidate subgraph size.}
We next vary the PCST input size $k_v=k_e$ while fixing $\lambda=0.001$. As shown in Table~\ref{tab:sens_pcst}, downstream QA Hit@1 increases from $74.51\%$ to $76.97\%$ as the candidate-subgraph size increases. This behavior is expected because a larger PCST subgraph provides a higher-recall candidate pool from which the learned retriever can select evidence.

\begin{table}[t]
\centering
\begin{tabular}{cc}
\hline
PCST input $k_v = k_e$ & Hit@1 (\%) \\
\hline
3  & 74.51 \\
5  & 75.43 \\
10 & 76.97 \\
\hline
\end{tabular}
\caption{Sensitivity analysis on the PCST candidate subgraph size on WebQSP, with $\lambda$ fixed to $0.001$.}
\label{tab:sens_pcst}
\end{table}

\section{Discussion of Answer-Aware or Answer-Related Methods and Additional Recent Baselines}
\label{app:additional-baselines}

Table~\ref{tab:appendix_baseline_comparison} summarizes several key distinctions between \algname\ and recent related baselines.
Here, in this comparison, \emph{LLM-based end-to-end} refers to a unified retrieve-then-generate pipeline that pairs a learnable \emph{LLM-based retriever} for graph evidence selection with a fine-tuned \emph{LLM-based generator} (i.e., the retriever itself is made a trainable LLM, rather than only fine-tuning the downstream generator).
\emph{Coverage-compactness} indicates whether the method explicitly aims to improve answer-relevant evidence coverage while suppressing irrelevant noise.
\emph{KGQA} indicates whether the method is directly applicable, as-is, to open-ended KGQA benchmarks such as WebQSP and CWQ.
\emph{Label-prediction GraphQA} indicates whether the method is directly applicable, as-is, to graph QA tasks whose outputs are discrete labels rather than entity answers, such as ExplaGraphs-style support/counter prediction.

\begin{table*}[t]
\centering
\begin{tabular}{lcccc}
\toprule
Method & LLM-based end-to-end & Coverage-compactness & KGQA & Label-prediction GraphQA \\
\midrule
\algname & \cmark & \cmark & \cmark & \cmark \\
\midrule
QA-GNN    & \xmark & \xmark & \xmark & \xmark \\
GNN-RAG   & \xmark & \xmark & \cmark & \xmark \\
\midrule
RoG       & \xmark & \xmark & \cmark & \xmark \\
CG-RAG    & \xmark & \xmark & \xmark & \xmark \\
SentGraph & \xmark & \xmark & \xmark & \xmark \\
\bottomrule
\end{tabular}
\caption{Comparison of \algname\ with answer-related and other recent graph retrieval methods discussed in Appendix~\ref{app:additional-baselines}.}
\label{tab:appendix_baseline_comparison}
\end{table*}

\paragraph{Answer-aware or answer-related methods.}
Some related methods do exploit answer-aware or answer-related signals, but they do so in task formulations that differ substantially from ours.

\paragraph{QA-GNN \cite{yasunaga2021qa}.}
QA-GNN is designed for multiple-choice QA rather than our open-ended GraphQA setting.
Given a question and an answer choice $a \in C$, it forms a QA context $[q;a]$, links entities mentioned in the question and answer choice to a KG, and performs joint LM+GNN reasoning over a constructed subgraph for that question-choice pair.
Therefore, QA-GNN is conditioned on \emph{candidate answer choices}, rather than using dataset gold answers to supervise a retriever.
Its main focus is answer-choice scoring in multiple-choice QA, not training a retriever for open-ended graph QA generation.
Accordingly, it does not satisfy our \emph{LLM-based end-to-end} criterion, does not explicitly optimize the \emph{coverage-compactness} trade-off, and is not directly applicable as-is to either open-ended KGQA benchmarks such as WebQSP/CWQ or label-prediction graph QA tasks such as ExplaGraphs.

\paragraph{GNN-RAG \cite{mavromatis2024gnn}.}
GNN-RAG targets KGQA.
It uses training question-answer pairs to supervise a GNN that scores answer candidates over a dense KG subgraph, and then retrieves the shortest paths connecting question entities to those candidates as context for a tuned LLM.
Thus, GNN-RAG does use answer supervision during training, but in the form of answer-candidate scoring and path retrieval for KGQA, rather than weighted pseudo-label supervision for graph evidence selection.
In particular, it does not contain a trainable \emph{LLM-based retriever}, nor does it explicitly optimize the \emph{coverage-compactness} trade-off.
It is directly applicable to KGQA, but not to label-prediction graph QA tasks such as ExplaGraphs, whose outputs are discrete labels rather than KG entities.

\paragraph{Other recent graph retrieval baselines.}
We also discuss several other recent graph retrieval methods that are useful points of comparison but differ substantially in task setting and retrieval target.

\paragraph{RoG \cite{luo2023reasoning}.}
RoG is a planning-retrieval-reasoning framework tailored to KGQA.
It first generates relation paths as plans, then retrieves valid reasoning paths from the KG and reasons over them with the same LLM.
Although RoG jointly trains its planning and reasoning modules, its retrieval step itself is plan-guided KG path retrieval rather than a trainable \emph{LLM-based retriever} under our definition above.
Moreover, RoG is designed for KGQA where answers are expected to be KG entities reachable via KG paths.
Therefore, it does not satisfy our \emph{LLM-based end-to-end} criterion, does not explicitly optimize the \emph{coverage-compactness} trade-off, and is not applicable as-is to label-prediction graph QA tasks such as ExplaGraphs.

\paragraph{CG-RAG \cite{hu2025cg}.}
CG-RAG targets research question answering over citation graphs.
Its retrieval module ranks relevant chunks by combining sparse and dense retrieval signals with graph encoding over contextual citation graphs, and the retrieved chunk subgraphs are then used for generation.
This differs from our setting in both graph type and retrieval unit: CG-RAG performs chunk-level retrieval over citation graphs, rather than node-level evidence selection over a given QA graph.
Accordingly, it neither provides an \emph{LLM-based end-to-end} retriever-generator pipeline in our sense nor explicitly optimizes \emph{coverage-compactness}, and it is not directly applicable as-is to either KGQA benchmarks such as WebQSP/CWQ or label-prediction graph QA tasks such as ExplaGraphs.

\paragraph{SentGraph \cite{liang2026sentgraph}.}
SentGraph is a sentence-level graph-based RAG framework for multi-hop QA over document collections.
It constructs hierarchical sentence graphs with explicit logical dependencies and performs graph-guided sentence retrieval and path expansion.
In contrast, our setting assumes a given QA graph whose nodes and edges already define the reasoning space, such as Freebase entities/relations in WebQSP/CWQ or commonsense concept graphs in ExplaGraphs.
Thus, SentGraph differs from our setting in graph granularity, retrieval target, and benchmark formulation.
It also does not provide an \emph{LLM-based end-to-end} retriever-generator pipeline in our sense, does not explicitly optimize \emph{coverage-compactness}, and is not directly applicable as-is to either KGQA benchmarks such as WebQSP/CWQ or label-prediction graph QA tasks such as ExplaGraphs.

Overall, the distinctive feature of \algname\ is not merely that it
leverages answer-related signals, but that it combines a unified
\emph{LLM-based end-to-end} retriever-generator pipeline with weighted
pseudo-label supervision, thereby addressing both the
\emph{architectural gap} and the \emph{retrieval-supervision gap} while
explicitly improving the \emph{coverage-compactness} trade-off.
Moreover, the framework applies to both open-ended KGQA benchmarks
(e.g., WebQSP and CWQ) and label-prediction GraphQA tasks
(e.g., ExplaGraphs).

\section{Concrete Prompt Templates and Graph Serialization}
\label{app:prompt-serialization}

In this section, we provide one concrete example of the prompts used in \algname. 
We show the exact retriever prompt and the generator prompt, and clarify how retrieved nodes and associated edges are serialized for answer generation.

\subsection{Retriever Prompt}
\label{app:retriever-prompt}

The retriever is prompted with the question together with the initial PCST candidate subgraph $\mathcal{S}$. 
Its task is to output nodes that are most useful for answering the question, and to output them as a single pipe-separated line.
Below we show one concrete retriever prompt example.

\begin{promptbox}{Retriever Prompt}
<s>[INST] <<SYS>>
You are an entity retriever for KGQA.
Your job is to SELECT CANDIDATE ENTITIES FROM THE SUBGRAPH that would help an answerer solve the question.
CRITICAL CONSTRAINTS:
- Return lowercase names only.
- Select the most relevant nodes from the subgraph for answering the question.

RETURN FORMAT (MANDATORY):
Your ENTIRE reply must be EXACTLY ONE SINGLE LINE like:
name1|name2|name3
If you output anything other than the single pipe-separated line, the answer is invalid.

Hard rules:
- Output EXACTLY one line; no greetings, no explanations, no extra words.
- The line must contain ONLY lowercase entity names separated by '|' (no labels like 'Output:').
- Keep spaces inside names; DO NOT add spaces around '|'.
- DO NOT output placeholders like A/B/C/unknown.

<</SYS>>

Question:
where did kennedy 's inaugural address take place

Subgraph (CSV between <DESC> tags):
<DESC>
node_id,node_attr
24,john f. kennedy
282,cold war
286,speech
287,m.05rrkys
295,anti-apartheid
352,m.04tnzwc
397,m.05r7b2w
428,john f. kennedy 1961 presidential inauguration
429,inaugural address of john f. kennedy
475,m.0b9zgjq
661,berlin wall
737,john fitzgerald kennedy memorial
771,john fitzgerald kennedy inaugural poem read by robert lee frost
967,united states presidential inauguration
1008,john f. kennedy speech
1119,apollo program
1282,the kennedy presidential press conferences
1335,1957 national book awards
1376,m.0ypz1mv
1458,1957 pulitzer prize
1701,jfk inauguration

src,edge_attr,dst
24,event.public_speaker.speeches_or_presentations,397
352,award.award_honor.award_winner,24
24,award.award_nominee.award_nominations,475
428,common.topic.image,1701
24,book.author.works_written,1282
24,music.artist.track,1008
967,time.recurring_event.instances,428
24,base.kwebbase.kwtopic.connections_from,771
737,symbols.namesake.named_after,24
1376,event.speech_or_presentation.type_or_format_of_presentation,286
287,event.speech_or_presentation.type_or_format_of_presentation,286
397,event.speech_or_presentation.type_or_format_of_presentation,286
1119,event.speech_topic.speeches_or_presentations_on_this_topic,1376
295,event.speech_topic.speeches_or_presentations_on_this_topic,287
661,event.speech_topic.speeches_or_presentations_on_this_topic,397
282,event.speech_topic.speeches_or_presentations_on_this_topic,397
428,base.inaugurations.inauguration.inaugural_address,429
397,event.speech_or_presentation.speech_topic,282
397,event.speech_or_presentation.speech_topic,661
24,base.inaugurations.inauguration_speaker.inauguration,428
475,award.award_nomination.ceremony,1335
352,award.award_honor.ceremony,1458
428,base.inaugurations.inauguration.president,24
287,event.speech_or_presentation.speech_topic,295
1376,event.speech_or_presentation.speech_topic,1119
</DESC>

Now output EXACTLY ONE SINGLE LOWERCASE LINE with '|' separators. Nothing else. [/INST]
\end{promptbox}

\subsection{Generator Prompt and Graph Serialization}
\label{app:generator-prompt}

For answer generation, we serialize the retrieved evidence in the format: 
nodes are first listed as a table with header \texttt{node\_id,node\_attr}, followed by edges as a table with header \texttt{src,edge\_attr,dst}.

The serialized node set contains two parts. 
First, nodes with regular numeric identifiers are retrieved nodes that appear in the initial PCST candidate subgraph $\mathcal{S}$. 
Second, nodes whose identifiers start with the prefix \texttt{na} (e.g., \texttt{na1}, \texttt{na2}) are retrieved nodes that do not appear in the initial $\mathcal{S}$. 
These additional nodes are retained in the generator context as isolated nodes. 
After listing the nodes, we include the associated edges $\hat{\mathcal{E}}$ that appear in $\mathcal{S}$. 
Therefore, the generator context consists of the retrieved nodes $\hat{\mathcal{V}}$ together with their associated edges $\hat{\mathcal{E}}$, and it is not required to form one connected subgraph.
Below we show one concrete generator prompt example.

\begin{promptbox}{Generator Prompt}

<s>[INST] <<SYS>>
You are a KGQA answer generator.
PRIMARY GOAL: output the FINAL ANSWER entity name(s) to the question.
Use the provided subgraph as STRONG HINTS. If some correct answers are not in subgraph but answers the question, you may also include them.
Return all names in lowercase.

RETURN FORMAT (MANDATORY):
Your ENTIRE reply must be EXACTLY ONE SINGLE LINE like:
name1|name2|name3
If you output anything other than the single pipe-separated line, the answer is invalid.

Guidelines:
- Output ONLY answers; exclude distractors and the subject itself unless the subject IS the answer.
- If multiple entities correctly answer the question, include them all separated by '|'. Otherwise, output a single name.
- No greetings, no explanations, no labels (e.g., 'Answer:').
- Keep spaces inside names; DO NOT add spaces around '|'.
- DO NOT output placeholders (A/B/C/unknown).

<</SYS>>

Question:
where did kennedy 's inaugural address take place

Subgraph (CSV between <DESC> tags):
<DESC>
node_id,node_attr
24,john f. kennedy
429,inaugural address of john f. kennedy
1008,john f. kennedy speech
1282,the kennedy presidential press conferences
1701,jfk inauguration
na1,washington d.c.
na2,capitol building
na3,united states
na4,1961 presidential inauguration
na5,united states capitol building

src,edge_attr,dst
428,common.topic.image,1701
24,book.author.works_written,1282
24,music.artist.track,1008
428,base.inaugurations.inauguration.inaugural_address,429
24,base.inaugurations.inauguration_speaker.inauguration,428
428,base.inaugurations.inauguration.president,24
</DESC>

Now output ONLY the final answer entity name(s) as EXACTLY ONE SINGLE LOWERCASE LINE with '|' separators. Nothing else. [/INST]
\end{promptbox}

\begin{figure*}[t]
\centering
\includegraphics[width=0.95\textwidth]{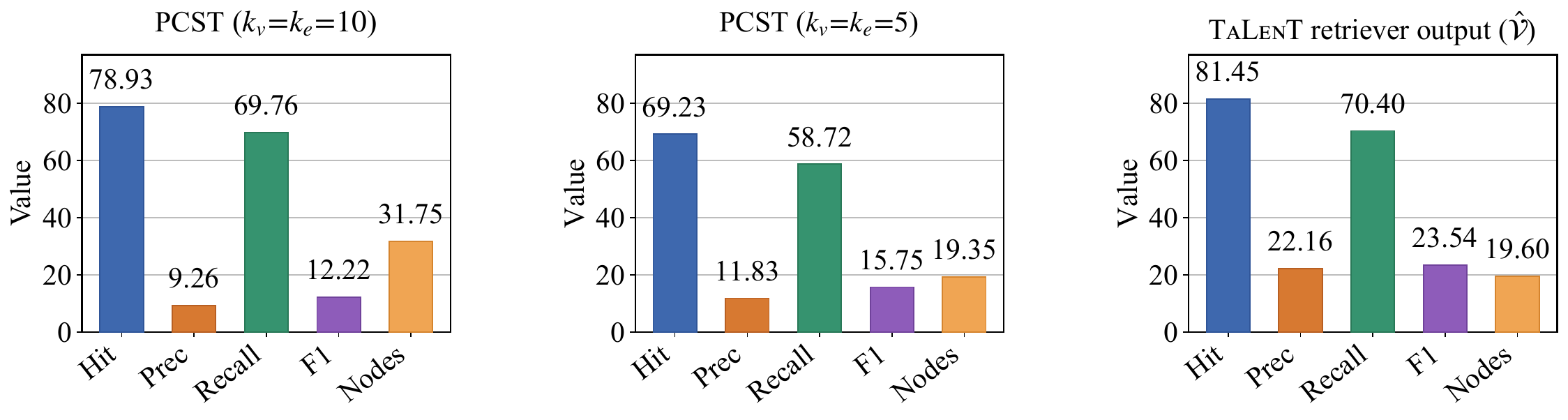}
\caption{WebQSP test-set retrieval results with Llama-3-8B (macro-averaged).
We measure how well PCST subgraphs and \algname's learned retriever output $\hat{\mathcal{V}}$ cover the gold answer set $\mathcal{V}^{\star}$.
We compare: (1) the PCST candidate subgraph with $k_v{=}k_e{=}10$ (the retriever input),
(2) a smaller PCST subgraph with $k_v{=}k_e{=}5$ (a size-matched reference),
and (3) the node set $\hat{\mathcal{V}}$ selected by \algname's learned retriever.
Our retriever maintains comparable gold-answer coverage with substantially fewer nodes than the original PCST candidate subgraph ($k_v{=}k_e{=}10$), while achieving higher coverage than the size-matched PCST subgraph ($k_v{=}k_e{=}5$).}
\label{fig:retr_efficiency_llama3}
\end{figure*}

\section{Groundedness and Hallucination Analysis of Out-of-PCST Retrieval}
\label{app:groundedness-hallucination}

Retrieving nodes beyond the initial PCST candidate subgraph can improve coverage, but it also raises a natural concern about groundedness: such nodes may introduce ungrounded evidence and potentially contribute to hallucinated answers. To assess this effect, we conduct an additional analysis on the WebQSP test set.

We use a practical \emph{hit-aware hallucination proxy}, since groundedness is ultimately relevant insofar as it affects the final answer. For each test instance, let $\mathcal{S}$ denote the initial PCST candidate subgraph with node set $\V_{\mathcal{S}}$, let $\hat{\mathcal{V}}$ denote the retrieved node set output by the retriever, and let $\mathcal{V}^{\star}$ denote the gold answer set. We define the hallucinated retrieved nodes as
\begin{equation}
\mathcal{H} =
\begin{cases}
\emptyset, & \text{if Hit@1} = 1,\\
\{v \in \hat{\mathcal{V}} \setminus \V_{\mathcal{S}} \mid v \notin \mathcal{V}^{\star}\}, & \text{if Hit@1} = 0.
\end{cases}
\end{equation}
That is, if the final prediction is correct, we do not count retrieved nodes as hallucinated under this proxy. If the final prediction is incorrect, we count as hallucinated those retrieved nodes that are outside the PCST candidate subgraph and do not match any gold answer.

We then compute the overall hallucination rate over the test set as
\begin{equation}
\text{Hallucination Rate}
=
\frac{\sum |\mathcal{H}|}{\sum |\hat{\mathcal{V}}|},
\end{equation}
where the sums aggregate over all test instances.

On the WebQSP test set, the resulting hallucination rate is $4.97\%$. This suggests that although the trained retriever can retrieve nodes beyond PCST, such retrieval introduces only a small fraction of ungrounded nodes overall. Combined with the results in Section 4.4 of the main paper, this indicates that out-of-PCST retrieval provides useful complementary coverage while maintaining a low level of hallucinated evidence.

\section{Backbone Ablation with Llama-3-8B}
\label{app:llama3_ablation}

We further evaluate \algname\ with Llama-3-8B on WebQSP, comparing against G-Retriever w/ LoRA, the strongest baseline in Table~\ref{tab:main_results}. As shown in Table~\ref{tab:llama3_main}, \algname\ achieves 83.05\% Hit@1, compared with 79.36\% for G-Retriever w/ LoRA, suggesting that our framework remains effective with a stronger LLM backbone. We use $\lambda=0.001$ for the learned retriever.

\begin{table}[t]
\centering
\caption{WebQSP downstream QA results with Llama-3-8B, comparing \algname\ against the strongest baseline G-Retriever w/ LoRA in Table~\ref{tab:main_results} (Hit@1, \%).}
\begin{tabular}{lc}
\toprule
\textbf{Method} & \textbf{Hit@1} \\
\midrule
G-Retriever w/ LoRA & 79.36 \\
\textbf{\algname} & \textbf{83.05} \\
\bottomrule
\end{tabular}
\label{tab:llama3_main}
\end{table}

We further examine the retrieval coverage-compactness tradeoff under Llama-3-8B, as shown in Figure~\ref{fig:retr_efficiency_llama3}.

\noindent\textbf{Compactness: comparable coverage with fewer nodes.}
Compared with the initial PCST candidate subgraph ($k_v{=}k_e{=}10$), \algname's retriever produces a more compact node set while maintaining comparable gold-answer coverage. Specifically, the average number of nodes decreases from 31.75 to 19.60, while Hit@1 increases from 78.93\% to 81.45\% and Recall from 69.76\% to 70.40\%. Precision also improves from 9.26\% to 22.16\%, and F1 from 12.22\% to 23.54\%.

\noindent\textbf{Coverage: better gold-answer coverage under a comparable node budget.}
We further compare against a smaller PCST subgraph with $k_v{=}k_e{=}5$, which contains 19.35 nodes on average, close to the 19.60 nodes retrieved by \algname. Under this comparable node budget, \algname\ improves Hit@1 from 69.23\% to 81.45\% and Recall from 58.72\% to 70.40\%. These results exhibit a similar coverage--compactness trend to the main experiments.

{\setlength{\tabcolsep}{3pt}
\begin{table}[t]
\centering
\caption{Breakdown of gold-answer coverage contributed by different parts of the learned retriever output on the WebQSP test set with Llama-3-8B ($\lambda=0.001$). All metrics are macro-averaged over examples and reported as percentages.}
\begin{tabular}{lcc}
\toprule
\textbf{Output subset} & \textbf{Hit@1} & \textbf{Recall} \\
\midrule
All retrieved nodes & 81.45 & 70.40 \\
Nodes within PCST & 76.90 & 67.35 \\
Nodes out of PCST & 8.29 & 4.28 \\
\bottomrule
\end{tabular}
\label{tab:retr_breakdown_llama3}
\end{table}
}

\noindent\textbf{Retrieval beyond PCST.}
Table~\ref{tab:retr_breakdown_llama3} further breaks down the gold-answer coverage of the learned retriever output. Nodes outside PCST alone achieve 8.29\% Hit@1 and 4.28\% Recall. Incorporating these nodes increases Hit@1 from 76.90\% to 81.45\% and Recall from 67.35\% to 70.40\%. This is consistent with the main-paper observations that the learned retriever can complement the PCST candidate subgraph with additional gold-answer coverage.

Overall, these results suggest that \algname\ maintains a favorable retrieval coverage--compactness tradeoff and strong downstream QA performance with Llama-3-8B.

\section{More Related Works}

\paragraph{Graphs and Large Language Models}
A growing body of work studies how to combine graph-structured information with LLMs \cite{li2023survey, wang2023can}.
Prior efforts span general graph modeling \cite{liu2023one, lei2023boosting}, multimodal architectures that incorporate graph-like structures \cite{yoon2023multimodal, li2023graphadapter}.
Representative applications include fundamental graph reasoning \cite{zhao2023graphtext}, node prediction \cite{sun2023large, chen2024exploring}, graph-level prediction \cite{qian2023can}.
A related line of work discusses GraphRAG / GraphQA, where the system selects graph evidence and conditions an LLM on graph context to answer questions \cite{logan2019barack,luo2025chatrule}.

\paragraph{Parameter-Efficient Fine-Tuning}
Parameter-efficient fine-tuning (PEFT) methods enable adapting large pretrained models with limited trainable parameters.
Representative techniques include prompt tuning \cite{lester2021power}, prefix tuning \cite{li2021prefix}, and low-rank adaptation such as LoRA \cite{hu2022lora}.
Related adapter-style methods (e.g., LLaMA-adapter \cite{zhang2023llama}) have also supported rapid adaptation and inspired multimodal instruction-following systems such as MiniGPT-4 \cite{zhu2023minigpt} and NExT-Chat \cite{wu2024next}.
PEFT has recently been explored in graph+LLM settings as well, including GraphLLM \cite{chai2025graphllm} and GraphToken \cite{perozzi2024letgraphtalkingencoding} for graph reasoning, and GNP \cite{tian2024graph} for knowledge-graph QA.

\end{document}